\documentclass[sigconf]{acmart}

\usepackage{booktabs} % For formal tables
\DeclareMathOperator{\E}{\mathbb{E}}

\setcopyright{acmcopyright}
\copyrightyear{2018} 
\acmYear{2018} 
\setcopyright{acmcopyright}
\acmConference[DLRS 2018]{3rd Workshop on Deep Learning for Recommender Systems}{October 6, 2018}{Vancouver, BC, Canada}
\acmBooktitle{3rd Workshop on Deep Learning for Recommender Systems (DLRS 2018), October 6, 2018, Vancouver, BC, Canada}
\acmPrice{15.00}
\acmDOI{10.1145/3270323.3270329}
\acmISBN{978-1-4503-6617-5/18/10}

\begin{document}
\title{Item Recommendation with Variational Autoencoders\\ and Heterogeneous Priors}

\author{Giannis Karamanolakis}
\affiliation{%
  \institution{Columbia University}
  \city{New York}
  \state{NY}
  \postcode{10025}
}
\email{gkaraman@cs.columbia.edu}

\author{Kevin Raji Cherian}
\affiliation{%
  \institution{Columbia University}
  \city{New York}
  \state{NY}
  \postcode{10025}
}
\email{kr2741@columbia.edu}

\author{Ananth Ravi Narayan}
\affiliation{%
  \institution{Columbia University}
  %\streetaddress{P.O. Box 1212}
  \city{New York}
  \state{NY}
  \postcode{10025}
}
\email{ar3792@columbia.edu}

\author{Jie Yuan}
\affiliation{%
  \institution{Columbia University}
  \city{New York}
  \state{NY}
  \postcode{10025}
}
\email{jyuan@cs.columbia.edu}

\author{Da Tang}
\affiliation{%
  \institution{Columbia University}
  \city{New York}
  \state{NY}
  \postcode{10025}
}
\email{datang@cs.columbia.edu}

\author{Tony Jebara}
\affiliation{%
  \institution{Columbia University, Netflix}
  \city{New York}
  \state{NY}
    \city{Los Gatos}
  \state{CA}
  \postcode{10025}
}
\email{jebara@cs.columbia.edu}

\renewcommand{\shortauthors}{G. Karamanolakis et al.}

\begin{abstract}
In recent years, Variational Autoencoders (VAEs) have been shown to be highly effective in both standard collaborative filtering applications and extensions such as incorporation of implicit feedback. 
We extend VAEs to collaborative filtering with side information, for instance when ratings are combined with explicit text feedback from the user. 
Instead of using a user-agnostic standard Gaussian prior, we incorporate user-dependent priors in the latent VAE space to encode users' preferences as functions of the review text. 
Taking into account both the rating and the text information to represent users in this multimodal latent space is promising to improve recommendation quality. 
Our proposed model is shown to outperform the existing VAE models for collaborative filtering (up to 29.41\% relative improvement in ranking metric) along with other baselines that incorporate both user ratings and text for item recommendation. 
\end{abstract}

\begin{CCSXML}
<ccs2012>
 <concept>
  <concept_id>10010520.10010553.10010562</concept_id>
  <concept_desc>Computer systems organization~Embedded systems</concept_desc>
  <concept_significance>500</concept_significance>
 </concept>
 <concept>
  <concept_id>10010520.10010575.10010755</concept_id>
  <concept_desc>Computer systems organization~Redundancy</concept_desc>
  <concept_significance>300</concept_significance>
 </concept>
 <concept>
  <concept_id>10010520.10010553.10010554</concept_id>
  <concept_desc>Computer systems organization~Robotics</concept_desc>
  <concept_significance>100</concept_significance>
 </concept>
 <concept>
  <concept_id>10003033.10003083.10003095</concept_id>
  <concept_desc>Networks~Network reliability</concept_desc>
  <concept_significance>100</concept_significance>
 </concept>
</ccs2012>
\end{CCSXML}

\ccsdesc[500]{Information Systems~Recommender systems}

\keywords{Item Recommendation, Variational Autoencoders, Deep Learning,  Probabilistic Modeling, Text Mining}

\maketitle
\newpage
\section{Introduction}
The ever-growing amount and diversity of user content available in online platforms has stimulated interest in robust recommender systems that effectively handle multimodal information. 
Deep learning models, including Variational Autoencoders (VAEs), have led to substantial progress in recommender systems over the last years~\cite{van2013deep,wang2014improving,wang2015collaborative,wu2016collaborative,covington2016deep,cheng2016wide,jannach2017recurrent,li2017collaborative,liang2018variational}, mainly because these models capture non-linear user-item relationships. 
In particular, extending VAEs using a multinomial likelihood~\cite{liang2018variational} has been shown to be highly effective for collaborative filtering with implicit feedback, as it generalizes linear factor models and is shown to outperform state-of-the-art baselines for item recommendation. 

While conventional VAEs have been shown to be effective for recommender systems, some important limitations remain. 
Indeed, it has been argued that the choice of a too simplistic prior like the standard Gaussian distribution could lead to poor latent representations~\cite{hoffman2016elbo}. Introducing richer priors in the VAE has attracted significant scientific interest over the last years~\cite{goyal2017nonparametric,nalisnick2017stick,nalisnick2016approximate,chen2016variational}. Another issue is that conventional VAEs for item recommendation do not consider other types of metadata available (e.g., text feedback from the user in addition to the rating) and this makes these models inferior to hybrid recommendation systems which combine collaborative filtering with content-based methods~\cite{wang2011collaborative,yang2011collaborative,mcauley2013hidden,chen2015recommender,smirnova2017contextual,musto2017multi,gopalan2014content}.
Addressing these two issues is a promising step towards improving recommender systems as it benefits from both the modeling capacity of deep learning models and the ability of hybrid models to handle heterogenous information. 

Towards this goal, we extend VAEs to collaborative filtering with side information, particularly in the case of online item reviews (e.g., Yelp reviews, IMDB movie reviews) where users provide text feedback in addition to a numeric rating.
We replace the user-agnostic standard Gaussian prior with heterogenous, user-dependent priors, which are estimated empirically as functions of the user's review text.
This modeling choice leads to user representations in a multimodal latent space encoding both user ratings and text.
As users tend to use the review text to justify their rating, harnessing this information is promising to improve recommendation performance.
To our knowledge this is the first approach to use heterogenous user-dependent priors in VAEs for recommendation. 
We show that our proposed method outperforms existing VAE models along with other hybrid baselines for item recommendation.

\begin{figure}[t]
    \centering
    \includegraphics[height = 4cm,width = 8cm]{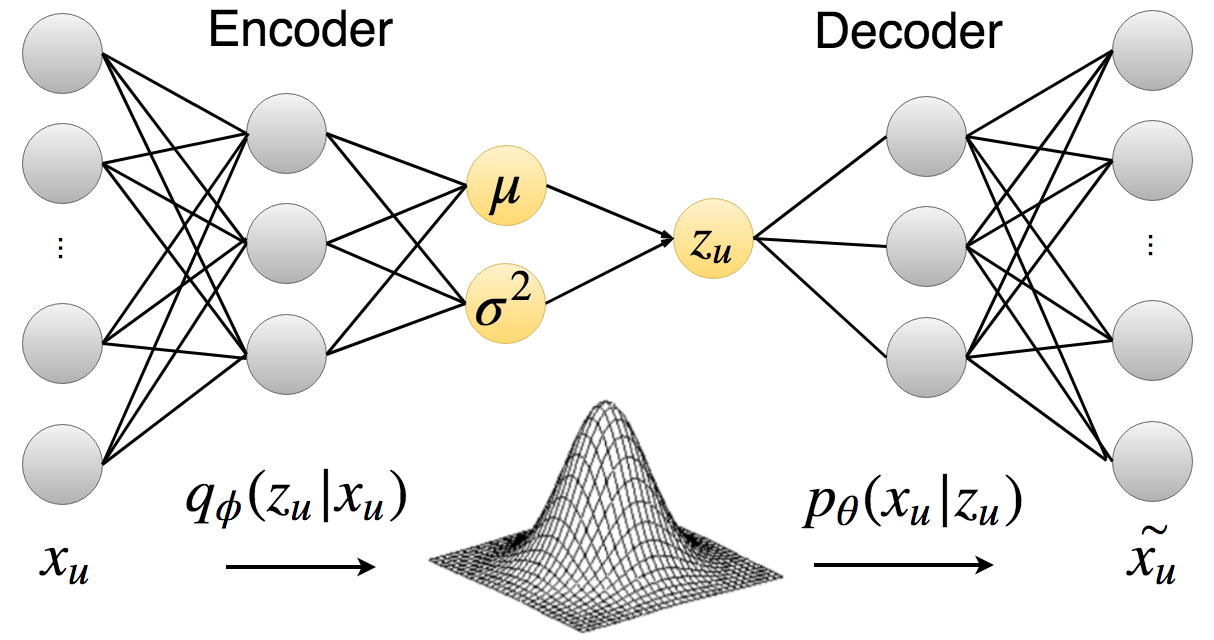}
   	\caption{Variational Autoencoders for collaborative filtering. For each user: (1) a bag-of-items vector $x_u$ is provided as input to the encoder; (2) a latent user vector $z_u$ is sampled from a Gaussian distribution with parameters specified by the encoder; (3) a new bag-of-items vector $\overset{\sim}{x_u}$ is reconstructed via the decoder.}
     \label{fig:vae}
   	%\vspace{-2mm}
\end{figure}

\section{VAE for Collaborative Filtering}
\label{sec:vaes}
Figure~\ref{fig:vae} represents VAEs for collaborative filtering. 
We define the variable $u \in {1,...,U}$ to index the users and $i \in {1,...,I}$ to index the items. The user-item matrix is defined as $\textbf{X} \in \mathbb{N}^{U \times I}$.
Each user $u$ is associated with a matrix row $x_u \in \mathbb{N}^I$, which is a ``bag-of-items'' vector (i.e., a bag-of-words vector over items).
In the case of online reviews for instance, 
$x_u$ encodes the ratings of user $u$, where items not rated by user $u$ are represented by zero values.

VAEs for collaborative filtering~\cite{liang2018variational} are probabilistic models that use the user-item matrix $\textbf{X}$ to obtain user representations in a latent low dimensional space. 
Next, we describe the generative process, the inference process, and the procedure followed to rank items for recommendation. 
\subsubsection{Generative model (decoder)}
\label{s:generative-model}
The generative process starts by drawing a latent representation $z_u \in \mathbb{R}^K$ for user $u$ from a Gaussian prior with zero mean and identity covariance matrix: 
\begin{equation}
\label{eq:standard-prior}
    z_u \sim \mathcal{N}(0,\mathbb{I}_K).
\end{equation}
Next, a non-linear function $f_{\theta}$ followed by the softmax function maps the user representation $z_u$ to a probability distribution over items $\pi(z_u) \in \Delta^{I-1}$ (($I-1$)-simplex):
\begin{equation}
    \pi(z_u) \propto \exp{\{f_{\theta}(z_u)}\}.
\end{equation}
The user's bag-of-items vector $x_u \in \mathbb{R}^I$ is assumed to be drawn from a multinomial distribution with success probability $\pi(z_u)$:
\begin{equation}
    x_u \sim Mult(N_u,\pi(z_u)),
\end{equation}
where $N_u = \sum_{i} x_{ui}$. The log-likelihood for $x_u$ is:
\begin{equation}
    \log p_{\theta}(x_u | z_u) = \sum_{i} x_{ui} \log{\pi_i(z_u)}.
\end{equation}
The multinomial distribution is well suited for modelling item-count data and is experimentally shown to outperform other types of likelihood functions for collaborative filtering \cite{liang2018variational}. 

\subsubsection{Inference model (encoder)}
During inference, the goal is to approximate the intractable posterior distribution $p_\theta(z_u|x_u)$. Using amortized variational inference \cite{gershman2014amortized}, the true posterior is approximated using a simpler (and tractable) variational distribution: 
\begin{equation}
    q_{\phi}(z_u) = \mathcal{N}(\mu_{\phi}(x_u),diag \{\sigma_{\phi}^2 (x_u)\}),
\end{equation}
where the parameters $\mu_{\phi}(x_u) \in \mathbb{R}^K$ and $\sigma_{\phi} (x_u) \in \mathbb{R}^K$ are computed by the inference model (encoder).

\subsubsection{Learning Procedure}
Under the variational inference setting, we compute the evidence lower bound (ELBO) for user $u$ as:

\begin{align}
\begin{split}
    \label{eq:ELBO}
    \mathcal{L}(x_u ; \theta, \phi) &\equiv \E_{q_{\phi}(z_u | x_u)}[\log{p_{\theta}(x_u|z_u)}] - KL(q_{\phi}(z_u|x_u) || p(z_u)) \\
    &\leq \log p(x_u;\theta).
\end{split}
\end{align}
The first term of the ELBO can be interpreted as the negative reconstruction error, while the second term (Kullback-Leibler divergence) can be interpreted as a regularization term forcing the variational posterior $q_{\phi}(z_u|x_u)$ to be near the prior $p(z_u)$.
It is common practice however \cite{liang2018variational,jang2016categorical,sonderby2016ladder,bowman2016generating,kim2018semi,yang2017improved} 
to introduce an additional parameter ($\beta$) to the ELBO:
\begin{align}
\begin{split}
    \label{eq:KL-annealing-ELBO}
    \mathcal{L_{\beta}}(x_u ; \theta, \phi) \equiv \E_{q_{\phi}(z_u | x_u)} &  [\log{p_{\theta}(x_u|z_u)}]\\
    &- \beta \cdot KL(q_{\phi}(z_u|x_u) || p(z_u)) .
\end{split}
\end{align}
 
Annealing $\beta$ over time from 0.0 to 1.0 is a successful approach that addresses the ``posterior-collapse'' phenomenon, where the variational posterior collapses to the prior \cite{bowman2016generating}. 
The VAE is trained by maximizing the ELBO with respect to parameters $\phi$ and $\theta$ (back-propagation is performed via the reparameterization trick \cite{kingma2013auto,rezende2014stochastic}).

\subsubsection{Item Ranking}
Given a test user $u$, the encoder ($q_{\phi}(z_u|x_u)$) infers the parameters $\mu_{\phi}$ and $\sigma_{\phi}^2$ given the bag-of-items vector $x_u$ and a latent user vector $z_u$ is sampled from a Gaussian distribution parameterized by $\mu_{\phi}$ and $\sigma_{\phi}^2$.
A new vector $\overset{\sim}{x_u}$ is reconstructed as a distribution over the items via the decoder ($p_{\theta}(x_u|z_u)$).
By optimizing Equation~\ref{eq:KL-annealing-ELBO}, the model is expected to assign more probability mass to the items that are more likely to be selected by a user. 
If this is true, then the model also performs well under a top-N ranking loss, which is often used to evaluate recommender systems. 

\section{Method}
We extend the VAE model described in Section~\ref{sec:vaes} to consider additional metadata of online reviews as well as the rating information. In this body of work we focus on the text provided by users in their reviews, but our approach could be easily extended for other types of metadata as well.

\subsection{User-dependent Priors}
\label{s:user-dependent-priors} 
Our proposed model extends the conventional VAE by replacing the standard prior of Equation~\ref{eq:standard-prior} with a user-dependent prior: 
\begin{equation}
\label{eq:user-depentent-prior}
    z_u \sim \mathcal{N}(t_u,\mathbb{S}_u),
\end{equation}
where $t_u \in \mathbb{R}^K$ and $\mathbb{S}_u \in  \mathbb{R}^{K\times K}$ are the mean vector and the covariance matrix respectively corresponding to user $u$. 
We show how to estimate the parameters of a user's distribution from his text in Section~\ref{s:encoding-user-preferences}. 
This modification to the generative model of the VAE allows each user to be modeled using a different prior distribution, enhancing in this way the expressitivity of the model.
Also, it is promising to address the ``posterior-collapse'' issue of VAEs~\cite{bowman2016generating} without necessarily using KL annealing strategies, because the user-dependent priors explicitly enforce diversity in the latent VAE space.
The extended ELBO becomes: 
\begin{align}
\begin{split}
    \label{eq:ELBO-with-user-dependent-priors}
    \mathcal{L_{\beta}}(x_u ; \theta, \phi, t_u,\mathbb{S}_u) \equiv& \E_{q_{\phi}(z_u | x_u)}[\log{p_{\theta}(x_u|z_u)}] \\
    &\quad- \beta \cdot KL(q_{\phi}(z_u|x_u) || p_u(z_u; t_u, \mathbb{S}_u)). 
\end{split}
\end{align}
This equation differs from Equation~\ref{eq:KL-annealing-ELBO}, as it regularizes the variational posterior $q_{\phi}(z_u|x_u)$ to be near the user prior $p_u(z_u; t_u, \mathbb{S}_u)$ for each user $u$. 
This modeling choice is simple and efficient because it yields a closed-form equation for the computation of the Kullback-Leibler divergence term of Equation~\ref{eq:ELBO-with-user-dependent-priors} and also it models each separate user using a different prior distribution. 
\subsection{Encoding user preferences from text}
\label{s:encoding-user-preferences}
Text from online reviews is a rich source of information, revealing user preferences towards specific aspects of the items. 
Modeling user preferences from the review text is of significant scientific interest~\cite{wang2011collaborative,yang2011collaborative,mcauley2013hidden,chen2015recommender,smirnova2017contextual,musto2017multi} as it can provide complementary information to that modeled by the traditional collaborative filtering systems. 
We present two different approaches for representing each user as a multivariate Gaussian probability distribution using the text of his reviews.

\subsubsection{Word Embeddings} 
Word embeddings have attracted substantial interest due to their ability to capture the semantic similarity between words \cite{bengio2003neural,collobert2008unified,mikolov2013distributed,pennington2014glove}. 
Pretrained word embeddings can be used to compute embeddings of phrases, sentences, paragraphs and documents \cite{arora2016simple,wieting2015towards,iyyer2015deep,chen2017efficient, le2014distributed} and quite interestingly, simple averaging methods over word embeddings outperform more complex methods~\cite{arora2016simple,wieting2015towards}. 
Here, we compute review embeddings as the average of pretrained word2vec embeddings \cite{mikolov2013efficient,mikolov2013exploiting,mikolov2013distributed}. 
Each user is modeled as a Gaussian distribution having a mean vector $t_u$ equal to the element-wise average of his review embeddings and a diagonal covariance matrix $S_u$ with diagonal values equal to the standard deviation of his review embeddings. 

\subsubsection{Topic Models}
Probabilistic topic models have been successfully used to capture global semantic coherency \cite{blei2009topic}. 
Latent Dirichlet Allocation (LDA) \cite{blei2003latent} is an unsupervised topic modeling approach, which we employ here. First, we train LDA on the set of the text reviews to learn the distributions of $k$ topics.
For each user we concatenate all of his reviews in one document and apply the learned LDA model to get a user representation as the distribution over the learned $k$ topics. 
\vspace{-1mm}
\section{Experimental Procedure}%\& Evaluation Procedure}
\subsection{Evaluation Datasets}
For evaluation, we use the Yelp Challenge Dataset\footnote{https://www.yelp.com/dataset} and the
IMDB corpus of movie reviews~\cite{diao2014jointly}. 
Table~\ref{tab:dataset-statistics} reports statistics of these datasets.
The Yelp Challenge Dataset contains customer reviews of local businesses. Each review is associated with a star rating ranging from $1$ to $5$ stars. We filter out reviews whose text is not written in English and businesses other than restaurants.  
We binarize the user item matrix using value $3$ as a threshold. 
To reduce sparsity, we filter out users providing less than 5 reviews and businesses that have been rated by less than 30 users (``Yelp cutoff'' in Table~\ref{tab:dataset-statistics}. 
The IMDB dataset contains movie reviews associated with ratings ranging from $1$ to $10$ stars. We binarize ratings using value $5$ as a threshold. We filter out users providing less than 5 reviews and movies that have been rated by less than 5 users (``IMDB cutoff'' in Table~\ref{tab:dataset-statistics}). 
\vspace{-2mm}
\subsection{Experimental Setting}
For comparison reasons, we follow the same experimental procedure as in~\cite{liang2018variational}. We split both datasets  into training (80\%), validation (10\%) and test (10\%) datasets. All training users' ratings are used for training the VAE. For validation and testing we use 80\% of a user's reviews to compute a hidden user representation $z_u$ and then use $z_u$ to predict a ranking over the $I$ items. The remaining 20\% of the user's reviews are used to evaluate the predicted item ranking. 

\begin{table}[t]
\centering
%\resizebox{\columnwidth}{!}{
\begin{tabular}{ c c c c c }
\hline
\textbf{Dataset}  & \textbf{\#users} & \textbf{\#items} & \textbf{\#ratings} & \textbf{sparsity}\\ 
\hline%\hline
Yelp & 930496 & 65536 &20000263 & 0.053e-3\% \\ %\hline
Yelp cutoff & 92208 & 13085 & 1257420 & 0.104\%  \\ %\hline
\hline
IMDB & 50331 & 21740 & 278907 & 0.025\% \\ %\hline
IMDB cutoff  & 8080 & 8357 & 167593&  0.248\% \\ %\hline
\hline
\end{tabular}%}
\caption{Dataset Statistics. ``sparsity'' refers to the percentage of non-empty entries in the user-item matrix.}
\label{tab:dataset-statistics}
\vspace{-4mm}
\end{table}

\begin{table*}[t]
\parbox{.45\linewidth}{
\centering
\resizebox{\columnwidth}{!}{
\begin{tabular}{ c c c c c}
\hline
\textbf{Model} & \textbf{Text Feat} & \textbf{NDCG@100} & \textbf{Recall@20} & \textbf{Recall@50}\\ %\hline
\hline
RAND &- & 0.006 & 0.002 & 0.005 \\ %\hline
MF &- & 0.066 & 0.070 & 0.071 \\ %\hline
Text-kNN & word2vec & 0.026 & 0.014 & 0.038\\ 
Mult-DAE &- & \textbf{0.178}& 0.217 & 0.309\\ %\hline %
Mult-VAE &- &0.147  &0.170  & 0.277 \\ %\hline
VAE-RP &- &0.148  & 0.193  & 0.298 \\ %\hline
VAE-TR  & word2vec & 0.149 & 0.212  & 0.290  \\% \hline
VAE-TR & LDA & 0.145 & 0.176  & 0.273 \\ %\hline 
\hline
VAE-HPrior & word2vec &\textbf{0.174} & 0.210  & \textbf{0.326} \\% \hline
VAE-HPrior & LDA & \textbf{0.174} & \textbf{0.220} & 0.322\\ %\hline
\hline
\end{tabular}}
\caption{Evaluation Results - IMDB Review Dataset. For VAE-TR we use $\gamma=0.01$. For MF we use $K=50$. The standard deviation of the scores is around $0.007$.}
\label{tab:imdb-results}
}
\hfill
\parbox{.45\linewidth}{
\centering
\resizebox{\columnwidth}{!}{
\begin{tabular}{ c c c c c}
\hline
\textbf{Model} & \textbf{Text Feat} & \textbf{NDCG@100} & \textbf{Recall@20} & \textbf{Recall@50}\\ %\hline
\hline
RAND &- & 0.001 & 0.001 & 0.002 \\ %\hline
MF &- & 0.070 & 0.010 & 0.030\\ %\hline
Text-kNN & word2vec & 0.003 & 0.002 & 0.006\\ %\hline
Mult-DAE &- &\textbf{0.121} &\textbf{0.152} & \textbf{0.252} \\ %\hline
Mult-VAE &- &0.104 & 0.123& 0.220  \\ %\hline
VAE-RP &- & 0.106& 0.129 & 0.225  \\ %\hline
VAE-TR & word2vec & 0.106 & 0.127 & 0.224  \\ %\hline
VAE-TR & LDA & 0.107  &0.127 & 0.225 \\%\hline % gamma 0.01
\hline
VAE-HPrior & word2vec & 0.114 & 0.137 & 0.238 \\ %\hline
VAE-HPrior & LDA &\textbf{0.119}  &\textbf{0.146}  &\textbf{0.247}  \\\hline
\end{tabular}}
\caption{Evaluation Results - Yelp Challenge Dataset. For VAE-TR we use $\gamma=0.01$. For MF we use $K=100$. The standard deviation of the scores is around $0.003$.}
\label{tab:yelp-results}
}
\vspace{-8mm}
\end{table*}
For the VAE, we use symmetric encoder and decoder models, which are parameterized by Multilayer Perceptrons. 
The input, hidden layer and output dimensionality of the encoder is $I$, 600 and 300 respectively, where $I$ is the total number of items, while for the decoder we use layers with dimensionality 300, 600, and $I$, respectively. 
The \textit{tanh} function is used as a non-linear activation between all layers except the output layer of the encoder. We apply dropout with probability 0.5 at the input layer, use a batch size of 500 users and update the model weights at training using the Adam optimizer for 50 epochs. 
For word embeddings we use the word2vec model, pretrained on Wikipedia articles with 300 dimensions. We apply z normalization to the user vectors. For topic models we train LDA on the Yelp and IMDB datasets separately using 300 topics. In both word2vec and LDA, each user is represented as a 300-dimensional multivariate Gaussian probability distribution.   
\label{s:modelparameters}
\vspace{-2mm}
\subsection{Evaluation Metrics}
For evaluation, we compute the same metrics as in~\cite{liang2018variational}.
In particular, for each user we compute the truncated Normalized Discounted Cumulative Gain at rank position 100 (NDCG@100), the Recall at position 20 (Recall@20) and Recall at position 50 (Recall@50). 
The final evaluation score is the average over all user scores. 
\vspace{-2mm}
\subsection{Baselines}
We compare our proposed method which we call VAE-HPrior (Heterogeneous Prior) with the following baselines:
\vspace{-2mm}
\subsubsection*{\textbf{Random (RAND)}} Ranking the items in random order.
\vspace{-2mm}
\subsubsection*{\textbf{Matrix Factorization (MF)}} The most widely used linear latent factor model. The number of latent dimensions ($K$) is fine-tuned on the validation set. We optimize using Stochastic Gradient Descent (SGD) with mini-batches of 1000 users. 
\subsubsection*{\textbf{kNN on Text Features (Text-kNN)}} A method based on text features. Both users and items are represented at the same vector space as the average of the word2vec embeddings of the reviews associated with them. Given a test user, the items are ranked according to their cosine similarities to the user representation.
\vspace{-1.5mm}
\subsubsection*{\textbf{VAE for Collaborative Filtering (Mult-VAE)}} The conventional VAEs for collaborative filtering~\cite{liang2018variational}. We use exactly the same architecture as in VAE-HPrior. VAE-HPrior differs from the Mult-VAE only in terms of adding heterogenous, user-dependent priors to the generative model (in VAE-HPrior) instead of a standard Gaussian prior (in Mult-VAE). 
\vspace{-1.5mm}
\subsubsection*{\textbf{DAE for Collaborative Filtering (Mult-DAE)}} A denoising autoencoder with multinomial likelihood, studied in \cite{liang2018variational}. Both the encoder and decoder are Multilayer Perceptrons with the same dimensionality as Mult-VAE and VAE-HPrior.  
\vspace{-1.5mm}
\subsubsection*{\textbf{VAE with Random User-Dependent Priors (VAE-RP)}} The same model as VAE-HPrior where instead of estimating the user-dependent priors from the text of the user's reviews, we use a random prior for each user. 
\vspace{-1.5mm}
\subsubsection*{\textbf{VAEs with Text Regularization (VAE-TR)}} 
A baseline using both the rating and textual information.
Instead of adding heterogenous priors, an extra regularization term is added to the extended ELBO of Equation~\ref{eq:KL-annealing-ELBO} that penalizes latent user representations $z_u$ that are distant from the corresponding user encodings $t_u$: 
\begin{align}
\begin{split}
    \mathcal{L_{\gamma}}(x_u ; \theta, \phi, t_u) =& \mathcal{L_{\beta}}(x_u ; \theta, \phi, t_u) - \gamma \cdot dist(z_u, t_u),
\end{split}
\end{align}
where  $dist(.)$ is the Euclidean distance metric.
VAE-TR introduces the parameter $\gamma$, which is fine-tuned on the validation set.
%\vspace{-1mm}
\section{Evaluation Results}
Table~\ref{tab:imdb-results} and Table~\ref{tab:yelp-results} report the evaluation results on the IMDB and Yelp datasets. 
On IMDB, VAE-HPrior's relative improvement in performance over Mult-VAE is 18.36\% for NDCG@100, 29.41\% for Recall@20, and 17.68\% for Recall@50. On Yelp, VAE-HPrior's relative improvement in performance over Mult-VAE is 14.42\% for NDCG@100, 18.69\% for Recall@20, and 12.27\% for Recall@50. 

The 29.41\% relative performance improvement indicates that the particular modeling choice of heterogenous user priors in VAE-HPrior is an effective extension to Mult-VAE. 
Also, as VAE-HPrior is parameterized by exactly the same encoder and decoder networks as Mult-VAE,
we could argue that taking into account the text of the user reviews leads to a richer latent VAE space which in fact improves recommendation performance. 
Compared to VAE-TR, which also combines the ratings with text features, VAE-HPrior is consistently better, achieving up to 25\% relative improvement in performance (for  Recall@20 on IMDB). 

We observe that Mult-DAE has comparable performance to VAE-HPrior on both datasets. 
Both models, however, perform significantly better than Mult-VAE on both datasets which indicates that using user-dependent priors instead of the too strong assumption of standard Gaussian priors is a way to make variational Bayesian approaches comparable to denoising approaches. 
Quite interestingly, VAE-RP has slightly better performance than Mult-VAE, indicating that VAEs could benefit from this simple modeling approach.
However, VAE-HPrior significantly outperforms VAE-RP, highlighting the importance of the text features extracted from user reviews for obtaining expressive user encodings. 

Using topic models for representing users is a slightly better approach compared to word embeddings.
Also, Text-kNN baseline leads to lower ranking scores than most baselines, which was expected, as this model doesn't consider historical user-item interactions. 
This underlines the importance of combining both historical user-item interactions and text features for getting better recommender systems.
\section{Conclusion and Future Work}
We extend Variational Autoencoders (VAEs) by replacing the standard Gaussian prior with user-dependent priors defined as functions of the review text. 
Using this approach, we map users into a latent vector space, which encodes both collaborative information and user preferences defined in review text. 
Our proposed model (VAE-HPrior) achieves up to 29.41\% relative improvement in a recall ranking metric compared to the conventional VAE and outperforms most of the recommendation baselines. 
While Mult-DAE performs comparably to VAE-HPrior in our experiments, we anticipate the advantages of VAE-HPrior to continue to grow with the increasing abundance of side-information. Users are increasingly more prone to share text information across social media with their networks rather than provide consistent numerical or quantitative ratings. 
Incorporating this information to more accurately model user preferences is promising to improve recommendation performance.  
In the future, we aim to experiment with more methods to incorporate side information to deep learning models in a way that gleans the best aspects of both variational Bayesian approaches and point estimates. 
Our method may also be improved by further exploration into aspect-based methods for the extraction of user preferences from text reviews. 
\newpage
\bibliographystyle{ACM-Reference-Format}
\bibliography{sample-bibliography}

\newpage
\end{document}